\pdfoutput=1

\documentclass[11pt]{article}

\usepackage[dvipsnames]{xcolor}
\usepackage{naacl2021}

\usepackage{times}
\usepackage{latexsym}

\usepackage[T1]{fontenc}

\usepackage[utf8]{inputenc}

\usepackage{microtype}

\usepackage{
  amsfonts, nicefrac, amsmath, bm, hyperref,
  soul, graphicx, enumitem, multirow, wrapfig, lipsum, booktabs, caption, subcaption, stfloats, listings,
  longtable, paralist
}
\usepackage[capitalise]{cleveref}

\newcommand{\ignore}[1]{}
\newcommand{\todo}[1]{}
\newcommand{\hoyle}[1]{}
\newcommand{\ana}[1]{}
\newcommand{\nascomment}[1]{}

\newcommand{\wrong}[1]{\textcolor{OrangeRed}{\textbf{#1}}}
\newcommand\sect[1]{\S\ref{#1}}

\title{Promoting Graph Awareness in Linearized Graph-to-Text Generation}

\author{Alexander Hoyle\textsuperscript{$\clubsuit$}\thanks{\, Work undertaken during an internship at AI2.} \hspace*{10mm} Ana Marasovi\'{c}\textsuperscript{$\dagger\Diamond$} \hspace*{10mm} Noah A. Smith\textsuperscript{$\dagger\Diamond$} \\ \\
  \textsuperscript{$\clubsuit$}Department of Computer Science, University of Maryland, College Park \\
    \textsuperscript{$\dagger$}Allen Institute for Artificial Intelligence\\
  \textsuperscript{$\Diamond$}Paul G. Allen School of Computer Science and Engineering, University of Washington\\
  \texttt{hoyle@umd.edu, \{anam,noah\}@allenai.org} \\}

\begin{document}
\maketitle

\maketitle
\begin{abstract}
Generating text from structured inputs, such as meaning representations or RDF triples, has often involved the use of specialized graph-encoding neural networks. However, recent applications of pretrained 
transformers to linearizations of graph inputs have yielded state-of-the-art generation results on graph-to-text tasks. Here, we explore the ability of these linearized models to encode local graph structures, in particular their invariance to the graph linearization strategy and their ability to reconstruct corrupted inputs. Our findings motivate solutions to enrich the quality of models' implicit graph encodings via scaffolding. Namely, we use graph-denoising objectives implemented in a multi-task text-to-text framework.
We find that these \emph{denoising scaffolds} lead to substantial improvements in downstream generation in low-resource settings.

\end{abstract}
\ignore{
    \emph{Legend for draft notes}
    \begin{compactitem}
        \item[] \hl{Writing to-dos}
        \item[] \hoyle{Comments/questions to readers}
        \item[] \todo{Additional coding/experiments} 
    \end{compactitem}

\noindent\ana{Title suggestions:}
\begin{compactitem}
    \item Robust Graph Linearizations for Graph-to-Text Generation 
    \item \nascomment{expressions if we want a ``cute'' title -- ``staying within the lines'' or ``farther down the line''}
\end{compactitem}
}

\section{Introduction}\label{sec:intro}


\begin{figure}[t]
  \centering
  \includegraphics[width=\columnwidth]{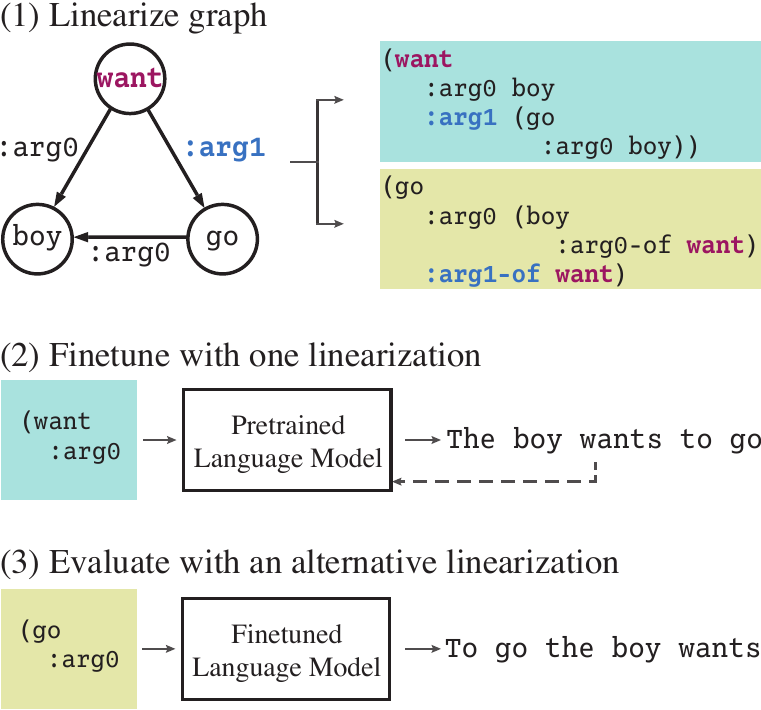}
  \caption{Diagram of our adversarial evaluation procedure for graph-to-text generation using pretrained language models (\sect{sec:linearization:shuffling}). (1) A graph can admit multiple possible linearizations. (2) Following standard practice, we train with a single linearization. (3) At evaluation time, we present the model with a meaning-preserving alternative.}
  \label{fig:motivation_fig}
\end{figure}


Parameter-rich pretrained transformer language models succeed at generating text that is prima facie fluent, but that closer inspection will often reveal to be semantically transgressive \cite{bisk-etal-2020-experience}.
Indeed, there is limited practical use for unconditional text generation: we expect language to relate to some identifiable, extrinsic meaning.
When a system communicates information to an individual in natural language, it will typically rely on a structured representation of that information.
Consequently, generating text that faithfully conveys structured data is an important goal in NLP, where inputs can take the form of tables \cite[ToTTo,][]{parikh-etal-2020-totto}, RDF triples \cite[e.g., WebNLG,][]{gardent-etal-2017-webnlg}, or Abstract Meaning Representations \cite[AMR,][]{flanigan-etal-2016-generation}.\looseness=-1

To accomplish this task, models have used neural architectures that explicitly encode graphs, such as graph neural networks \cite[GNNs,][]{Kipf2017SemiSupervisedCW} and graph transformers, in order to accurately capture the structural properties of the input graph \cite[][to name a few]{Zhu2019-wl,zhao-etal-2020-bridging,wang-etal-2020-amr}.
As an alternative to constraining a model architecture with a graph structure, another line of work \emph{linearizes} a graph into a string (Figure \ref{fig:graph-examples}) and trains a sequence-to-sequence model from scratch \cite{pourdamghani-etal-2016-generating,konstas-etal-2017-neural,Vinyals2015GrammarAA}. Initially, this approach was outperformed by graph-based encoders, but such models have recently seen their generation performance \textit{far surpassed} by pretrained transformer language models (LMs) finetuned on pairs of linearized graphs and their corresponding surface realizations \cite[][henceforth termed \emph{pretrained linearized models}]{mager-etal-2020-gpt,Kale2020-vd,Harkous2020-my,Ribeiro2020-mp}. Moreover, both automated and human assessments indicate that text generated with LMs retains meaning at least as well as graph-encoding baselines \cite{mager-etal-2020-gpt}.

This is not the sole product of pretrained models' general language knowledge: \citet{mager-etal-2020-gpt}, using a GPT-2-based \cite{radford2019language} model, report that ablating structural graph information (e.g., edges) in the linearized representation notably degrades generation performance, particularly in AMR-to-text tasks. \nascomment{I think I'm missing something here; how does the ablation in the preceding sentence lead to the conclusion in the next one?} The remarkable performance of pretrained linearized models is intriguing: explicit representation of the input graph by way of the model architecture appears to be well-substituted by simply writing the graph as a linear sequence.\looseness=-1

In this work, we further investigate the extent to which pretrained models can leverage linearized graph inputs.
Focusing on AMR graphs and sets of RDF triples in English-language datasets, we structure our investigation by first testing whether models' encodings are invariant to the linearization strategy---the way in which a graph is traversed and encoded when producing the linearized 
representation (see Figure  \ref{fig:motivation_fig}).  
We discover that generation suffers under adversarial permutations of the linearization, and embrace 
a simple-but-effective training strategy to mitigate this problem: adversarial training \cite{Goodfellow2015ExplainingAH}.
Motivated by this finding, we 
encourage more faithful encodings of graph structure via \ignore{unsupervised \nascomment{this word, unsupervised, might raise some hackles---it uses annotated data so is it really worth calling it unsupervised?}} denoising objectives in the more complex AMR setting. This multi-task \emph{scaffolding} \cite{swayamdipta-etal-2018-syntactic} reveals that straightforward masking of the graph input is sufficient to improve generation quality in low resource settings.\looseness=-1

Moreover, when treating this denoising performance as a proxy for the quality of models' implicit graph encoding, we find that it explains the semantic fidelity of the resulting generation better than reasonable alternatives, suggesting possibilities for future evaluation metrics.

We organize our investigation around two research questions:
\begin{compactitem}
    \item[\textbf{RQ1}] To what extent are pretrained linearized models invariant to graph linearization strategy? (\sect{sec:linearization})\looseness=-1
    \item[\textbf{RQ2}] Does encouraging pretrained linearized models’ implicit graph representation lead to better generation? (\sect{sec:scaffolding})\looseness=-1
\end{compactitem}


\section{Background: Graph-to-Text Generation}

In a graph-to-text setting, we transduce graph inputs $g$ to their corresponding surface realization $y = \langle y_1, \ldots, y_N\rangle$ via a parameterized probabilsitic 
model $p_\theta(\cdot)$.
In linearized models specifically, the graph $g$ is first mapped to text by way of a (usually deterministic) linearization function $x = l(g)$, where $p_\theta(\cdot)$ is an off-the-shelf sequence-to-sequence model.
This leads to the likelihood objective: $p_\theta(y \mid g) = \prod_{i=1}^N p_\theta(y_i \mid x, y_{1:i-1})$. 
\label{eq:gen_loss}
\ignore{\nascomment{rewrote left side as conditional prob., reordered stuff on right side of conditional bar on right to reflect linearization order; might want to replace comma there with some concatenation symbol}}
When $p_\theta(\cdot)$ is a left-to-right (autoregressive) pretrained transformer, generation quality far exceeds architectures with encoders specifically engineered to encode graphs \cite{mager-etal-2020-gpt,Kale2020-vd,Harkous2020-my,Ribeiro2020-mp}.\looseness=-1

\paragraph{Graph-to-Text Generation Datasets} 
\begin{table}[t]
        \centering
        \begin{tabular}{l r r r}
                \toprule
                 & \textbf{$N$} & \textbf{Dev. ppl.} & \textbf{Avg. edges}  \\
                \midrule
                \textbf{LDC2017T10} & 36k & 21.1 & 11.4 \\
                \textbf{WebNLG} & 18k & 9.2 & 3.0 \\
                \bottomrule
        \end{tabular}
        \caption{Dataset statistics. Perplexity estimated on the development set with GPT-2 \cite{radford2019language} fine-tuned on the training data using default hyperparameters in the \texttt{transformers} library \citep{Wolf2019HuggingFacesTS}.}
        \label{tab:datasets}
\end{table}

We explore two  datasets for generation from a graph structure to English text. 

Abstract Meaning Representation \cite[AMR,][]{banarescu-etal-2013-abstract} is a formalism intended to represent the propositional meaning of 
utterances---``who is doing what to whom''---using graphs that have minimal dependence on the surface form. AMR graphs are directed and acyclic with a single ``top'' node \cite{goodman-2020-penman}. They can be represented as either a graph, a tree, or sets of triples \cite{van2017neural}. For our data, we use the AMR 2.0 release (LDC2017T10),\footnote{\url{catalog.ldc.upenn.edu/LDC2017T10}} both because it spans a varied set of domains and styles, and because of its extensive use in prior work.

A simpler graph-to-text problem involves converting a set of RDF triples to natural text realizations of the information contained in the set, exemplified by the WebNLG dataset \cite{gardent-etal-2017-webnlg}. WebNLG pulls information from an existing knowledge base \cite[DBPedia,][]{mendes-etal-2012-dbpedia} for a specific subset of 15 categories (e.g., ``astronaut''). To generate the paired sentences, crowdworkers verbalize individual triples. Then, for examples consisting of multiple triples, they merge already-annotated sentences and apply minimal changes (leading to reduced sentence complexity relative to AMR, see perplexity scores in \cref{tab:datasets}). There can be multiple surface realizations per input.\looseness=-1

\paragraph{Models} To study pretrained linearized models' invariance to graph linearization, we use T5 \cite{Raffel2019-zl}, an \textit{encoder-decoder} transformer \cite{vaswani2017attention} that has led to state-of-the-art generation on AMR (specifically, LDC2017T10) and WebNLG \citep{Kale2020-vd,Ribeiro2020-mp}.

We modify the T5 implementation from the \texttt{transformers} library \cite{Wolf2019HuggingFacesTS}.\footnote{We use T5-Base for WebNLG and T5-Large for AMR, finding that the larger model did not benefit the WebNLG task.}
We use the Adafactor optimizer \cite{Shazeer2018AdafactorAL} with a learning rate of 0.0001, selected from the set $\{0.001, 0.0001, 3\times 10^{-5}, 1\times 10^{-5}, 1\times 10^{-6}\}$ after tuning on 1000 training examples across five random seeds.\footnote{Less extensive experiments with the full dataset indicated the same optimal setting, although in general it is relatively robust to learning rate.} We set the batch size to 6 and train until development set BLEU has not improved for 10 epochs. During decoding, we use a beam size of 10 for WebNLG and 5 for AMR.


\paragraph{Evaluation Measures} As a primary metric, we evaluate generated text using BLEU \cite{papineni-etal-2002-bleu}, calculated with \texttt{SacreBLEU} \cite{post-2018-call}. Despite its limitations in generation settings, BLEU still generally accords with rankings of models, either by human evaluations or by alternate metrics \cite{manning-etal-2020-human}.
We also evaluate our scaffolding models (\sect{sec:scaffolding}) using BertScore \cite{Zhang2020BERTScoreET}, which measures token similarity with contextual embeddings, permitting a more nuanced measure of semantic similarity. Lastly, we use the $\mathcal{M}$ portion of the $\mathcal{MF}$-score \cite{opitz2020towards}, which measures how well the source AMR graph can be reconstructed from the generated target sentence using an off-the-shelf parser. Unlike BLEU, which applies corpus-wide, this metric provides a best-guess at sentence-level accuracy.\looseness=-1

\section{RQ1: Robustness to Permutation of Graph Linearization}\label{sec:linearization}
\begin{figure*}[ht]
\centering
\begin{subfigure}[b]{0.3\textwidth}
     \centering
\begin{lstlisting}
(a / and
  :op1 (d / dream-01
      :ARG1 (f / film
          :ARG0-of (d2 / disturb-01))
      :ARG2-of (|*\bfseries r / resemble-01*|
            |*\bfseries :ARG1 a2*|))
  :op2 (a2 / and
      :op1 (f2 / fascinate-01
           :ARG0 f)
      :op2 d2))
\end{lstlisting}
     \caption{Canonical}
     \label{fig:graph-examples:canon}
\end{subfigure}
\hfill
\begin{subfigure}[b]{0.3\textwidth}
     \centering
\begin{lstlisting}
(a / and
  :op1 (d / dream-01
      :ARG2-of (r / resemble-01)
      :ARG1 (f / film
          :ARG0-of (f2 / fascinate-01)
          :ARG0-of d2))
  :op2 (|*\bfseries a2 / and*|
      :op2 (d2 / disturb-01)
      :op1 f2
      |*\bfseries :ARG1-of r*|))
\end{lstlisting}
     \caption{Reconfigured}
     \label{fig:graph-examples:rec}
\end{subfigure}
\hfill
\begin{subfigure}[b]{0.3\textwidth}
     \centering
\begin{lstlisting}
(|*\bfseries r / resemble-01*|
  :ARG2 (d / dream-01
      :op1-of (a / and
            :op2 a2)
      :ARG1 (f / film))
  |*\bfseries :ARG1 (a2 / and *|
       :op1 (f2 / fascinate-01
           :ARG0 f)
       :op2 (d2 / disturb-01
           :ARG0 f)))
\end{lstlisting}
     \caption{Randomized}
     \label{fig:graph-examples:rand}
\end{subfigure}
\hfill
\caption{Three PENMAN-based linearizations of AMR graphs corresponding to the sentence, ``The film is a dream and, like a dream, is both fascinating and disturbing.'' Note that the \textbf{bolded} relation in the graph, \texttt{(resemble-01 :ARG1 and)}, is represented differently depending on the linearization.
}
\label{fig:graph-examples}
\end{figure*}
                      








In this section, we explore the extent to which pretrained linearized models are invariant to the particular method used to linearize the input graph.
Motivated by the strong graph-to-text performance of these models, we ask: do they implicitly develop a robust internal encoding of the input graph?
Whereas a GNN-based model has an architecture designed for graph representation (e.g., information flows between adjacent nodes in a message-passing update), a linearized model must infer how connections are specified in a sequence during training.

If linearized models do form a representation, then the their estimates of the target sentence should be invariant to an alternative linearization of the same graph, so long as the original linearization is in principle recoverable from this alternative. If a model meets this criterion, we call it \textbf{linearization-invariant}.\looseness=-1
\ignore{ Why is this property  desirable? If we imagine the likely scenarios in which a graph-to-text model might be applied---say, a voice assistant---humans will not provide the meaning representations. Instead, they will be the output of some automated system. \nascomment{need more here.  do we think that automated systems will have highly variable output?  or that it can't be canonicalized somehow?  I'm not sure I believe this ...} \ana{I'm also struggling with this example: the system that produces the meaning representation will be a part of developers' ``meaning-to-text pipeline'' and developers will make sure there is no domain mismatch.}}

\subsection{Experimental Setup}
\label{sec:exp_setup}
To better understand models' graph-encoding behavior, we experiment with adversarial linearization strategies in two graph-to-text settings.

\paragraph{Permutations of AMR-Graph Linearizations}\label{sec:linearization:amr}

Standard AMR corpora are linearized as spanning trees over the graphs in \textsc{penman} notation (\citealt{matthiessen1991text}, see \cref{fig:graph-examples:canon}). 
In the present work, we also linearize graphs using \textsc{penman}, doing so for several reasons: (1) it is sufficiently flexible to accommodate significant changes to the linearization, discussed below; (2) it is more concise than sets of directed triples, both reducing training time and ensuring that inputs fit in the transformer context window; (3) the format leads to superior generation over reasonable alternatives, e.g., DFS traversal paths \cite{mager-etal-2020-gpt}.\looseness=-1

We will refer to the human-created linearizations in AMR corpora as {\textsc{canonical}}, since annotators  follow a standardized process. There is evidence that this format, in particular the relative ordering of edge types, leaks information about the associated sentence order \cite{konstas-etal-2017-neural}. We speculate that overparametrized models may overfit to such correlations rather than develop robust implicit graph encodings, since it has been repeatedly reported that large models use dataset shortcuts \cite[among others]{jia-liang-2017-adversarial, gururangan-etal-2018-annotation, geva-etal-2019-modeling}.

As an alternative linearization, \citet{goodman-2020-penman} defines the {\textsc{reconfigure}} operation as creating a tree from an AMR graph, where order information from the canonical linearization is ignored, except for the top node (e.g., \texttt{and} in \cref{fig:graph-examples:canon,fig:graph-examples:rec}). Although it is not a labeled element in the graph, the top node conveys structural information about the sentence---for instance, it is often the main verb. Reconfiguration can include reversals of edge labels (e.g., \texttt{ARG0} to \texttt{ARG0-of}), therefore constituting a substantive change to the linearization.

We also experiment with a more drastic restructuring of the graph, where we construct a tree from a {\textsc{randomized}} triple set alone, disregarding all order information from the canonical format (\cref{fig:graph-examples:rand}). Since it remains a valid traversal of the graph, in principle a model should be able to use this information to construct the surface sentence.

We parse, reconfigure, and randomize graphs using the \texttt{Penman} library \citep{goodman-2020-penman},\footnote{\url{github.com/goodmami/penman}} then replace variable names with their references and remove word sense information, following \citet{ribeiro-etal-2019-enhancing}.

\paragraph{Permutations of RDF-Triple Linearizations}\label{sec:linearization:rdf}

We follow the procedure of \citet{Ribeiro2020-mp} to form our standard linearization: we prepend a special token to each element of the triple, and separate triples with another dedicated token. For the output sentence ``Ned is the father of Rod and Todd,'' we would have:
\par\nobreak\vspace{-1em}{\small
\begin{align*}
     \textbf{In:}\ &(\text{Ned}\ \textit{fatherOf}\ \text{Rod}),\,(\text{Ned}\ \textit{fatherOf}\ \text{Todd}) \\
     \textbf{Out:}\ &\texttt{<rel> <S> Ned <V> father of <O> Rod}\\
     &\texttt{<rel> <S> Ned <V> father of <O> Todd }
\end{align*}
}%
For our adversarial permutation, we {\textsc{randomize}} the ordering of the relations.

\paragraph{Encouraging Robustness to Linearization} We train additional models with the goal of encouraging an agnosticism to graph linearization strategy. We adopt an adversarial training approach \cite{Goodfellow2015ExplainingAH}, and alter the graph linearization presented to the model at each epoch. We argue that this scheme ought to reduce any model dependence on the human-derived annotation.

\begin{table*}[ht]
        \centering
        \resizebox{\textwidth}{!}{
            \begin{tabular}{l r r r c r r}
                    \toprule
                                          & \multicolumn{3}{c}{\textbf{AMR (LDC2017T10)}} & &  \multicolumn{2}{c}{\textbf{WebNLG} (seen / unseen)} \\
                    \textit{Training linearization$\rightarrow$}  & {\textsc{Canonical}} &  {\textsc{Reconfigured}} & {\textsc{Randomized}} &  &  {\textsc{Canonical}} & {\textsc{Randomized}} \\
                    \midrule
                    \textit{Eval. linearization$\downarrow$}  &        &        &       & &                & \\
                    {\textsc{Canonical}}    & 43.52  &  43.08 & 40.90 & & 62.56 / 44.73  & 62.55 / 45.09\\
                    {\textsc{Reconfigured}} & 33.27  &  41.13 & 40.33 & &                &  \\
                    {\textsc{Randomized}}   & 22.89  &  31.00 & 39.80 & & 54.00 / 39.23  & 59.40 / 42.35\\
                    \midrule
                    \textit{GNNs}                       & --     &  --    & --    & &  --            & -- \\
                    \citet{wang-etal-2020-amr}  & 28.8  &  --    & --    & &  --            & -- \\
                    \citet{zhao-etal-2020-bridging}     & --     &  --    & --    & & 64.42 / 38.23  & -- \\
                    \bottomrule
            \end{tabular}
        }
        \caption{BLEU under different linearizations, using \textsc{T5-large} (AMR, development set\todo{update}) and \textsc{T5-base} (WebNLG, for both ``seen'' and ``unseen'' test sets).}
        \label{tab:results_shuffled}
\end{table*}

\subsection{Robustness Results}\label{sec:linearization:shuffling}

For both tasks, we train the model on the canonical linearization, then evaluate on the various linearizations described in \cref{sec:linearization:amr,sec:linearization:rdf}.

\paragraph{Impact of Adversarial Linearizations} The \textsc{canonical} columns of \cref{tab:results_shuffled} show results for models trained on that linearization, then evaluated on permuted graph linearizations. We note a strong negative impact in models' generation capacity for both tasks, with a starker decrease for the AMR data. These results suggest that pretrained linearized models are not linearization-invariant, failing to learn robust implicit graph representations, even in the case of the much simpler WebNLG data.\looseness=-1

The remaining columns of \cref{tab:results_shuffled} show that our straightforward adversarial training technique improves robustness, with only minor cost to generation performance. This is the case even with the more drastic \textsc{randomized} AMR linearization.
Moreover, it only incurs a minor impact on training time---for AMR, the \textsc{canonical}, \textsc{reconfigure}, and \textsc{randomize} variants attain 40 BLEU at 2, 3, and 5 epochs, respectively. \todo{(similar results hold for WebNLG)} \todo{[Evaluate a version trained with randomized + canonical mixture to get best of both worlds?]}

Given that elements of canonical annotations are known to correlate with the target sentence order \cite{konstas-etal-2017-neural}, we do not find it surprising that the models trained \emph{and} evaluated on the permuted linearizations show decreased performance.
However, it is meaningful that the canonical linearization at evaluation time still leads to the best results, even for models trained with the randomized inputs---these models did not learn to associate the canonical ordering signal with the input graph. One possible explanation is that the earlier pretraining induces a sensitivity to input token order that persists despite the adversarial fine-tuning, but the behavior merits further exploration.

\section{RQ2: Better Implicit Graph Encodings with Text-to-Text Scaffolding}\label{sec:scaffolding}
The positive results of our adversarial training procedure (\sect{sec:linearization:shuffling}) suggest that pretrained linearized models can form a robust internal graph representation, even though they rely on linearized inputs. Under substantively different linearizations, models retain the ability to generate accurately (even the \textsc{randomize} model outperforms best-in-class graph transformers; \citealt{wang-etal-2020-amr}).

Prior work, involving both GNNs and pretrained linearized models, has explored various ways of improving models' sensitivity to the structure of the input graph.
To better maintain fidelity to the graph, previous graph-to-text methods incorporate additional loss terms, specialized architectures, or generation-time ranking to influence the semantic accuracy of generation:  ranking outputs by the correctness of the AMR parse \cite{mager-etal-2020-gpt,Harkous2020-my}, jointly ``back-parsing'' graphs when decoding \cite{bai-etal-2020-online}, or using distinct  components to model different graph traversals \cite{ribeiro-etal-2019-enhancing}.

\ignore{Expanding on these efforts\ana{If above are indeed pretrained linearized models, we need to replace ``For the linearized case'' with something else.}, is it possible that inducing more accurate internal graph encodings leads to better generation?}
These efforts suggest that explicitly accounting for graph structure can assist generation. Can we expand on this idea, and improve generation quality by inducing more robust internal graph representations?
To answer this question, we propose secondary objectives designed to promote graph ``awareness.'' In addition to the above graph-to-text approaches, we also draw inspiration from \emph{denoising} methods used in language model pretraining \cite{Raffel2019-zl,lewis-etal-2020-bart}, as well as syntactic \emph{scaffolds} that support semantic tasks with an auxiliary syntax-dependent loss \cite{swayamdipta-etal-2018-syntactic}. Intermediate auxiliary pretraining has been repeatedly shown to be successful in other contexts \cite{Phang2018SentenceEO, ijcai2019-0249, gururangan-etal-2020-dont}.

\subsection{Experimental Setup}

In particular, we propose unsupervised graph-denoising tasks that we train alongside AMR-to-text generation, following the multi-task setup of \citet{Raffel2019-zl}. For each batch, we either optimize the likelihood in \cref{eq:gen_loss} or one of the objectives described below.\footnote{Per-task batches proved marginally better than mixing within a batch. The scaffolding task probability is a hyperparameter, which we set to 0.5.}


\paragraph{Masked Graph Modeling} When training transformers to have wide-ranging natural language capabilities, unsupervised denoising objectives like masked language modeling have proved extremely successful \cite{devlin-etal-2019-bert,Raffel2019-zl}. We argue that a similar principle ought to apply to graph understanding, and therefore apply masking directly to linearized graphs. 

In masked language modeling, each word token is masked with probability 15\%.  Here, we mask different sets of tokens, depending on the experimental condition, always setting the probability such that 15\% of \emph{all} tokens will be masked. Specifically, we mask: all tokens in the linearized graph, the graph components alone (edge labels and parentheses), and the semantic nodes. We also experiment with standard masking of the surface sentence, which mirrors the unsupervised domain-adapted pretraining employed by \citet{Ribeiro2020-mp}.\footnote{We use MASS-style masking \cite{Song2019MASSMS} for the tokens, rather than the span-replacing of T5, as it performed somewhat better.} For example, when masking components alone:
\par\nobreak\vspace{-1em}{\small 
\begin{align*}
    \textbf{orig}\ &\texttt{(   stupefy :ARG1 ( we )\ \ \ )}\\
    \textbf{in}\   &\texttt{(   stupefy <M>\ \ \ ( we\ <M> )}\\
    \textbf{out}\  &\textit{\ original text}
\end{align*}
}%
Graph masking can also be performed on any of the linearization variants defined in \cref{sec:linearization:amr}.\footnote{We restrict ourselves to the {\textsc{reconfigure}} setting given that early results showed little difference from {\textsc{randomize}}.}

\paragraph{Graph Reordering} Building on our findings from \cref{sec:linearization:shuffling}, we introduce a \textit{reordering} objective. Specifically, we provide the model with a \textsc{reconfigured} or \textsc{randomized} linearization, then task the model with reconstructing the canonical version.
We suspect that learning this mapping requires that the model captures 
the graph structure better, 
leading to superior graph-to-text generation.
Unlike the joint re-generation approach of \citet{mager-etal-2020-gpt}, where the input graph is copied alongside the target text, our method both requires a nontrivial encoding of the graph and has the effect of augmenting the data (due to the nondeterministic reconfiguration).\footnote{Simultaneously generating the surface text and reordering to the canonical linearization did not improve results.}

\nascomment{in tables, consider typesetting ``randomized'' and ``reconfigured'' with textsc like in main text?}

\ignore{
\paragraph{Composing and Decomposing Triples} As discussed in  \cref{sec:exp_setup}, \textsc{penman}-style notation is only one means of linearizing the graph. We also use linearizations more similar to that of the WebNLG data, namely, lists of ordered triples. \ana{This is a bit confusing because this is a new linearization that's not introduced in Section 3. Can you introduce it here a bit more formally? Also, add a comment why we anticipate this objective will lead to better capturing of graph, as it is done for graph reordering.} We include objectives to generate from a set of triples to a \textsc{penman}-linearized graph, as well as the reverse.
}


\subsection{Scaffolding Results}\label{sec:scaffolding:results}
\begin{table}[t]
        \centering
        \small{
                \begin{tabular}{l r}
                        \toprule
                         & BLEU  \\
                        \midrule
                        Baseline & 24.33 (0.94)   \\
                        Sentence masking (MLM)  & 27.73 (1.29)   \\
                        \textit{Graph Masking} &   \\
                        \quad All tokens & 28.48 (0.90)   \\
                        \quad Components & 28.49 (0.48)   \\
                        \quad Nodes      & \textbf{29.56} (1.05)   \\
                        \quad \textit{w/ \textsc{reconfigured} input} &   \\
                        \qquad All tokens & \textbf{29.41} (0.90)   \\
                        \qquad Components & 28.34 (0.58)   \\
                        \qquad Nodes      & 28.77 (0.80)   \\
                        \textit{Reordering to canonical} &   \\
                        \quad From \textsc{reconfigured} & 28.27 (0.90)   \\
                        \quad From \textsc{randomized} & 28.29 (0.91)   \\
                        \ignore{
                            \textit{Triples} &   \\
                            \quad Triples-to-\textsc{penman} & 27.15 (1.51)   \\
                            \quad Penman-to-triples & 28.09 (1.86)   \\
                            \quad Triples-to-text  & 25.45 (1.08)   \\
                        }
                        \bottomrule
                \end{tabular}
        }
        \caption{Development set BLEU across scaffolding objectives and baselines, trained on 1000-example subsets of the AMR dataset (LDC2017T10). Mean (s.d.) over 5 seeds.}\label{tab:results_scaffolding:n_1000}
\end{table}

\begin{table}[bhtp]
        \centering
        \resizebox{\columnwidth}{!}{
            \begin{tabular}{l r r r}
                    \toprule
                        & BLEU & BS & $\mathcal{M}$ \\
                    \midrule
                    \citet{bai-etal-2020-online}   & 34.19 &  - & - \\
                    \citet{Ribeiro2020-mp}         & 45.80 &  - & - \\
                    \midrule 
                    Baseline                         & 44.51 (0.48) & 77.40 (0.36) & 76.53 (0.19) \\
                    \emph{Scaffolding}             & & & \\
                    \ Mask nodes                     & 45.14 (0.23) & 77.75 (0.13) & 76.52 (0.14) \\
                    \ \textsc{Recon}., mask all    & 44.89 (0.39) & 77.56 (0.26) & 76.54 (0.20)  \\
                    \ Reorder from \textsc{recon}. & 44.86 (0.19) & 77.62 (0.16) & 76.34 (0.17)  \\
                    \bottomrule
            \end{tabular}
        }
        \caption{Test-set results of scaffolding objectives and baselines trained on the \textit{full} AMR dataset (LDC2017T10). \citet{bai-etal-2020-online} is a state-of-the-art graph transformer. \citet{Ribeiro2020-mp} finetunes \textsc{T5-large}, which we re-implement as our baseline model.
        BS is BertScore \cite{Zhang2020BERTScoreET}, and $\mathcal{M}$ is the meaning component of the $\mathcal{MF}$-score \cite{opitz2020towards}. Mean (s.d.) over 5 seeds.}\label{tab:results_scaffolding:amr}
\end{table}

\begin{figure}[t!]
  \centering
  \includegraphics[scale=.55]{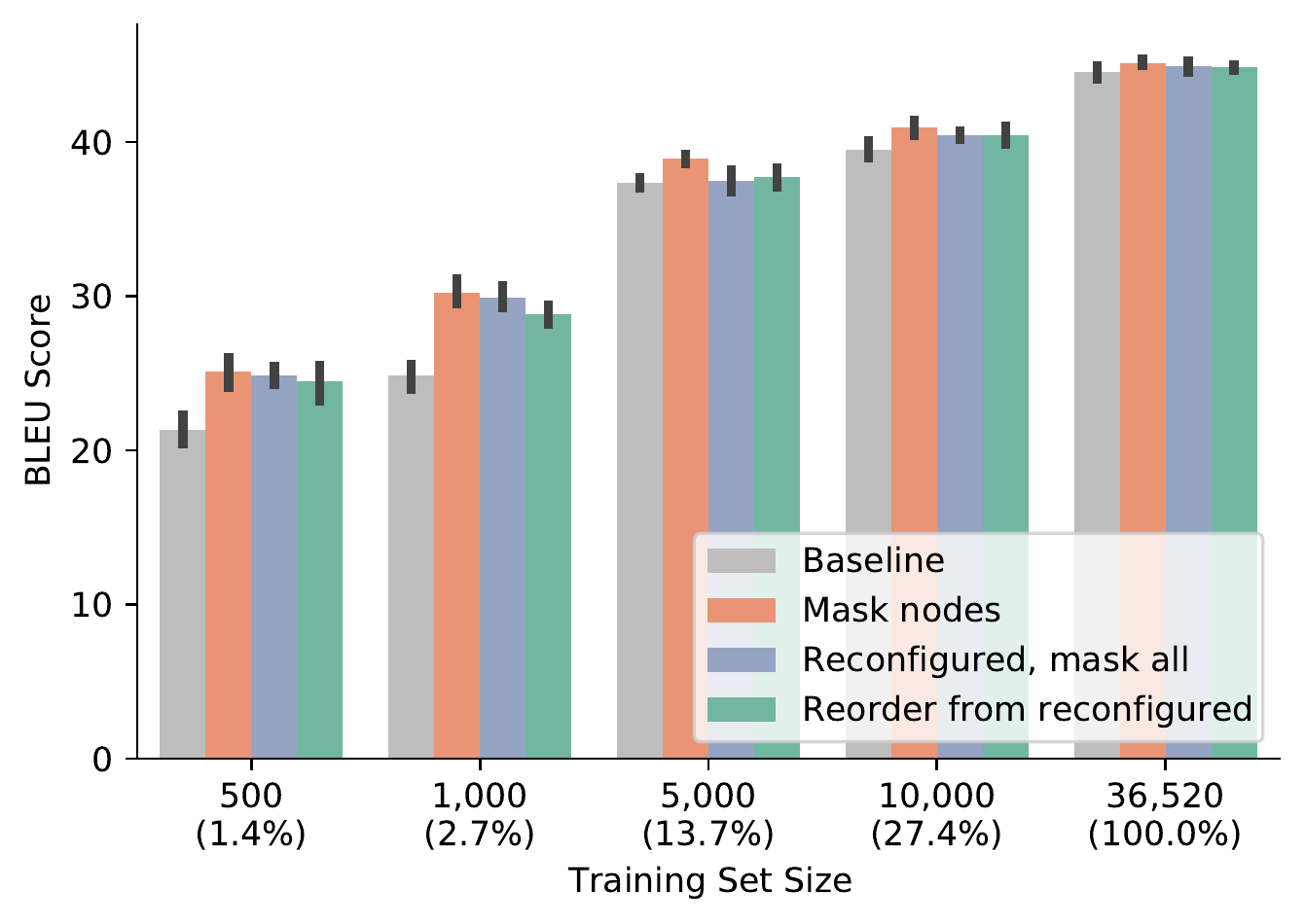}
  \caption{Test set BLEU on the AMR dataset (LDC2017T10) under different amounts of training data for selected scaffolding objectives (over 5 seeds).}
  \label{fig:limited_data}
\end{figure}
We find that, overall, denoising objectives drive substantial \todo{significance tests} improvements over the baseline when training on the reduced $n=1000$ dataset (\cref{tab:results_scaffolding:n_1000}).
In fact, using less than 3\% of the full data produces results that exceed that of state-of-the-art GNN models from a year prior to this writing \cite[BLEU 27.37,][]{ribeiro-etal-2019-enhancing}.
Moreover, the results suggest that focusing on the graph representation itself is most important: standard sentence masking (i.e., MLM-style) is less beneficial than graph masking, although it still outperforms the baseline.
Surprisingly, the various graph-masking objectives perform similarly to one another---there is little benefit to  more complex strategies that specifically account for the graph structure.

While the increased generation quality from the graph-denoising methods is not drastic relative to the MLM case, we contextualize our gains by noting that other ways of promoting greater graph awareness yield similar improvements in absolute terms---and come at the cost of greater model complexity or generation time. For instance, the use of two graph representations in \citet{ribeiro-etal-2019-enhancing} achieve a roughly 1-BLEU increase over the use of one alone.\looseness=-1

Based on the findings from the $n=1000$ setting (\cref{tab:results_scaffolding:n_1000}), we select three of the best-performing scaffolding objectives---\emph{mask nodes}, \emph{reconfigure \& mask all tokens}, and \emph{reorder from reconfigured}---and train them at $n \in \{ 500, 1000, 5000, 10000, N\}$.  Results are shown in  \cref{fig:limited_data}. At $n=5000$, representing 14\% of the data, the impact of scaffolding is no longer strong across all objectives.  When evaluating on the full dataset, the difference is minor (\cref{tab:results_scaffolding:amr}). For both BLEU and BertScore, we observe slight improvement over the baseline on average for the \emph{mask nodes} case, but it is within a standard deviation of the baseline (estimated over 5 seeds). $\mathcal{M}$-score does not vary between models, but it is also not yet established for fine-grained model selection.
It appears that the increased size of the data supplants the need for scaffolding losses: the sheer diversity of the source graphs encourages a graph-reasoning ability sufficient to generate accurate sentences.
Of course, in a realistic application,
hundreds or thousands of training examples are more attainable than tens of thousands. That such straightforward methods can yield strong gains is extremely promising for future work in low-resource graph-to-text generation.\looseness=-1

\begin{table*}[tp]
  \centering
  {\footnotesize
    \begin{tabular}{l | p{0.9\textwidth}}
      \toprule
      \textbf{Target} & \textit{Both Norway and Sweden have been spared violent terror acts but authorities in both countries have voiced concern about terrorists or terror financiers operating out of Scandinavia.}\\
      \textbf{Baseline} & \wrong{Norwegian and Swedish authorities have spared Norway and Sweden} from violent acts of terror but have voiced concern about terrorists or financiers of terror operating out of Scandinavia.\\
      \textbf{Ours} & Norway and Sweden have been spared terror acts of violence but Norwegian and Swedish authorities have voiced concern about terrorists or financiers of terror operating out of Scandinavia.\\
      \midrule
      \textbf{Target} & \textit{The 30-day simple yield fell to an average 8.19\% from 8.22\%; the 30-day compound yield slid to an average 8.53\% from 8.56\%.}\\
      \textbf{Baseline} & The simple 30 day yield \wrong{fell to 8.22 percent from 8.19 percent} on average and the compound 30 day yield \wrong{slid to 8.56 percent from 8.53} percent on average.\\
      \textbf{Ours} & Simple 30 day yields fell from 8.22 to an average 8.19\% and compound 30 day yields slid from 8.56 to an average 8.53\%.\\
      \midrule
      \textbf{Target} & \textit{Many young Saudi radicals have crossed the long and porous border between the Kingdom and Iraq and joined up with Sunni Muslim insurgents there.}\\
      \textbf{Baseline} & Many young Saudi radicals have crossed the porous border \wrong{from Iraq to the Kingdom} and joined up with Sunni Islamic insurgents there.\\
      \textbf{Ours} & Many young Saudi radicals have crossed the porous \wrong{long-term border} with Iraq and joined up with Sunni Islamic insurgents there.\\
      \bottomrule
    \end{tabular}}
  \caption{Selected predictions from the baseline and a model using the \textit{reordering-from-reconfigured} scaffold (trained on the full data).}
  \label{tab:generation_examples}
\end{table*}

\paragraph{Qualitative Analysis} In a manual analysis of 100 random model predictions, we generally observe broad agreement between the model trained with the \emph{reordering-from-reconfigured} scaffold and the baseline (73\% agreement in fidelity), both trained with the full dataset. However, in three cases, the baseline model fails to capture the order of arguments (e.g., ``from y to x'' when ``from x to y'' is correct), whereas the scaffolded model remains true to the graph (see \cref{tab:generation_examples}; we did not note instances of the reverse case). While we fail to note ``hallucinations''---material information that is not contained in the graph input---both models occasionally drop modifiers (e.g., adjectives or adverbs). Finally, a common error in both models is word-sense confusion (see the third row in Tab.~\ref{tab:generation_examples}, where ``long [in length]'' is substituted with ``long [in duration]''). This is likely due to the removal of word-sense suffixes during preprocessing to avoid sparsity issues  ($\texttt{long-03}\to\texttt{long}$). While currently standard practice, a system aiming to achieve perfect fidelity would require this data. \todo{[Currently, these results aren't based on the 'best' model, since it was  done earlier. Will need to clarify]}

\subsection{Encoding Graphs and Generation Performance}
\begin{figure}[t!]
  \centering
  \includegraphics[scale=.65]{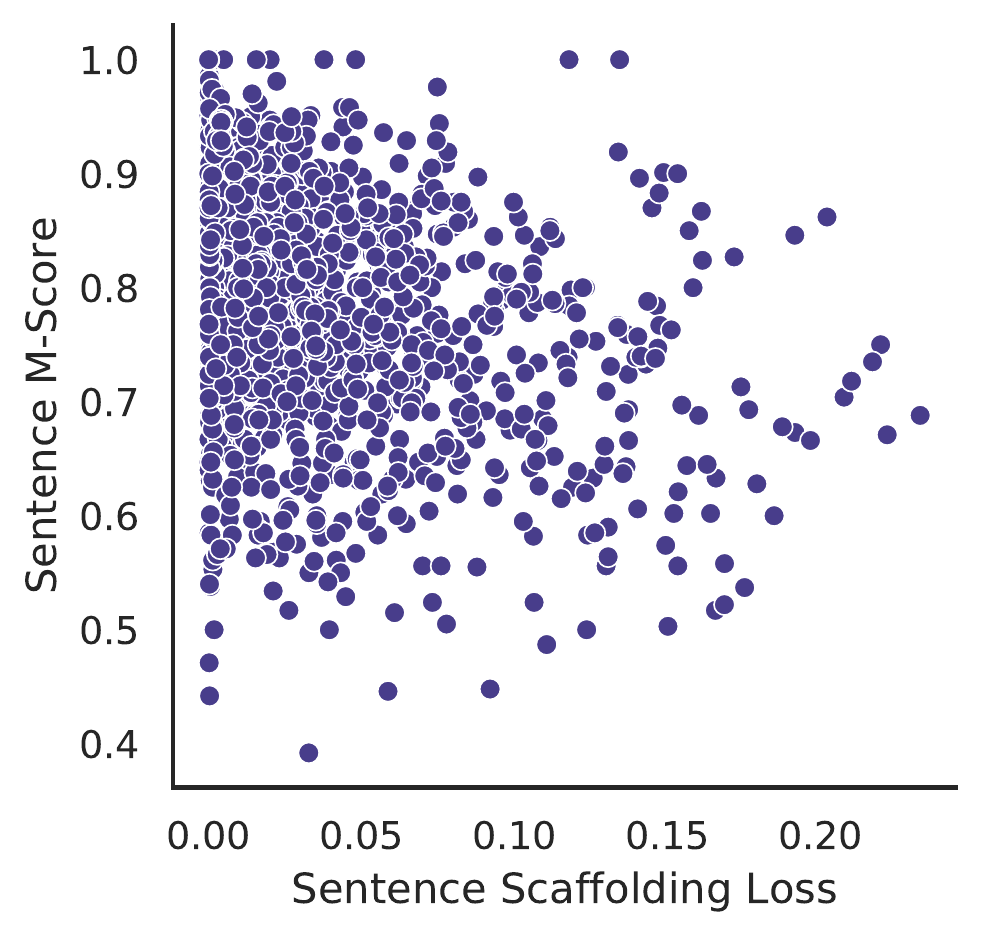}
  \caption{Sentence-level scaffolding loss and $\mathcal{M}$-score on the validation set, using a model trained with the \textit{reordering-from-reconfigured} scaffold. $\mathcal{M}$-score is a measure of the generated sentence's semantic fidelity, and the scaffolding loss is a proxy for the graph encoding accuracy.}
  \label{fig:loss_v_smatch}
\end{figure}

The results of Section \ref{sec:scaffolding:results} show that the denoising scaffolds impact generation performance. 
If we consider the sentence-level scaffolding loss as a proxy for the quality of its implicit graph encoding, can it help explain generation fidelity?\ignore{\ana{I'm missing a motivation that led to raising this question after Section 4.2.}}
In order to determine this relationship, we quantify generation accuracy using the $\mathcal{M}$ component of the $\mathcal{MF}$-score \cite{opitz2020towards}. It is calculated by first using an off-the-shelf parser to create an AMR graph from the generated target sentence, then by measuring the overlap with the gold source AMR (from 0 to 1). \ignore{To what extent can a model's generation fidelity be explained by the quality of its implicit graph encoding?}
As seen in \cref{fig:loss_v_smatch}, there is a substantial negative relationship (Pearson's $\rho=-0.35^*$) between these two variables, measured using outputs from the model trained with the \emph{reordering-from-reconfigured} scaffold on the full data.

To fully operationalize the above question, we estimate a linear regression on the $\mathcal{M}$ score of predicted sentences from the validation set. As covariates, we include the above (logged) scaffolding loss, in addition to other metrics that have a significant independent correlation with generation quality. In particular, we use sentence-BLEU, the number of edges in the graph, graph re-entrancies, words in the target sentence, and the (also logged) sentence generation loss.\footnote{We eliminate outliers consisting of the bottom 0.5\% of target lengths and $\mathcal{M}$-scores and the top 0.5\% of the losses.}

\begin{table}[t]
        \centering
        \small{
                \begin{tabular}{l r}
                        \toprule
                        $X$ & $\beta$ \\
                        \midrule
                        \textit{Intercept} & 0.7590*\\
                        Scaffolding loss (log) & -0.0094*\\
                        Generation loss (log) & -0.0088* \\
                        $\text{BLEU} / 100$  & 0.0628*\\
                        Words in target & -0.0021* \\
                        \midrule 
                        BIC & -2378\\
                        Adj. $R^2$ & 0.267 \\
                        \bottomrule
                \end{tabular}
        }
        \caption{OLS regression results on validation sentence $\mathcal{M}$-score, a measure of semantic fidelity. Model trained with the \emph{reordering-from-reconfigured} scaffold. *Significance at $p<0.001$.}
        \label{tab:results_regression}
\end{table}

We use the Bayesian information criterion (BIC) to select the model from all possible combinations of the above covariates. We find that the preferred model with $p$ covariates, $p=1\ldots6$, includes the reordering loss in all but one case ($p=2$), suggesting its validity as an indicator of graph fidelity above and beyond other alternatives. As seen in \cref{tab:results_regression}, it has a significant negative relationship with the $\mathcal{M}$ score, larger than that of the comparably-scaled generation loss. These results indicate that the reordering loss captures important information about the quality of the graph encoding.\looseness=-1

\ana{Make captions Table 5, Figure 3, and Table 6 more self-contained. A reader who reads only captions should somewhat understand what the experiments are about.}

\ignore{\ana{Should we provide evaluation of our models with the reordering loss?}}

\section{Related Work}\label{sec:prior-work}
\paragraph{Pretrained transformers for Graph-to-Text Generation} \citet{mager-etal-2020-gpt} condition GPT-2 \cite{radford2019language} on a linearized AMR graph, then fine-tune on the corresponding surface representation text. Later work using  
transformers has also found success on both AMR-to-text and data-to-text tasks \cite{Kale2020-vd,Harkous2020-my,Ribeiro2020-mp}. To our knowledge, across a diverse set of tasks and automated\footnote{Human evaluation has been less thorough, although \citet{mager-etal-2020-gpt} report improved human judgments on AMR-to-text generation. We note similar results in our own experiments.} metrics, a pretrained transformer of sufficient capacity will always outperform a specialized GNN, often by a large margin. 
\citet{Ribeiro2020-mp}, following \citealt{gururangan-etal-2020-dont}, further pretrain on additional in-domain data, using both supervised (silver AMR parses to text) and unsupervised (denoising target text) objectives.

\paragraph{Graph-Dependent Losses} \citet{mager-etal-2020-gpt} use various heuristics to improve fidelity. During training, they regenerate the input graph, and in inference, they parse generations and rank their consistency with the original graph. \citet{Harkous2020-my} instead rank with a trained classifier, and introduce additional ``state embeddings'' to help indicate the ordering of graph components.
The encoder-decoder methods cited in the previous paragraph eschew these approaches and nonetheless perform better. In preliminary replications of the Mager et al. experiments with T5, we find that joint re-generation leads to no improvement and moreover that the longer output sequences increase training time. Experimenting with other graph-sensitive embeddings is a valuable direction for future work.

\paragraph{Graph Linearization} Other work also studies linearizations for AMR-to-text settings. As opposed to our efforts, the focus is not on enriching or measuring models' graph encoding, but instead on determining what elements of linearization (e.g., parentheses and edge labels) are necessary for generation.\looseness=-1

Closest to our work is \citet{konstas-etal-2017-neural}, who experiment with alternative graph traversals by randomizing the edge type order (less drastic than either \textsc{reconfigure} or \textsc{randomize}) with an LSTM-based model. Rather than randomizing at each epoch, as in our approach, they employ a \emph{consistent} random ordering for each example during training,\todo{[If we mention this, may need to establish that it matters]} and do not evaluate models across different linearizations. The results help establish that LSTMs can be made agnostic to ordering, but fail to measure the extent to which models overfit to the training order (\cref{sec:linearization:shuffling}).

\citet{Ribeiro2020-mp} report paired training and evaluation shuffling results (as in \cref{tab:results_shuffled}), but they ignore parentheses, only reodering node labels. Hence, their results cannot establish models' graph-encoding ability, instead revealing that node order is informative of word order, corroborating findings in \citet{konstas-etal-2017-neural}.
Both works, along with \citet{mager-etal-2020-gpt}, run ablations by removing parenthetical markers, finding that graph structure is necessary for strong generation.

Finally, \citet{kedzie-mckeown-2020-controllable}, appearing contemporaneously to our work, seek to control the output generation by manipulating the input linearization order, using a randomization similar to ours as an ``uncontrolled'' baseline. Given their focus on task-oriented dialogue planning, which uses simpler meaning representations and sentences than the AMR dataset used here (i.e., shallower graphs and limited domains), we view their work as complementary to our own.


\section{Conclusion}

In this work, we explore the graph-encoding ability of pretrained transformers through the lens of graph-to-text generation that relies on linearized graph inputs. First, we determine the extent to which these models are invariant to the method by which graphs are linearized, finding that models trained on the fixed, canonical linearizations fail to generalize to meaning-preserving alternatives. We rectify this shortcoming by training models on linearizations corresponding to alternative random traversals of the graph.
Following prior work that has used graph-aware losses to improve generation quality, we then explore ways of improving models' sensitivity to the input graphs. Motivated by the success of denoising objectives in other text-to-text settings, we encourage robust internal graph encodings through additional scaffolding losses. Although scaffolding leads to tepid improvements in generation quality when training data is plentiful, it yields substantial gains in low-resource settings.\todo{Check with final  results}

\ana{Add links to all references.}

\bibliography{anthology,refs,paperpile}
\bibliographystyle{acl_natbib}



\end{document}